\documentclass[manuscript,screen]{acmart}
\acmJournal{TIST}
\usepackage{hyperref}
\usepackage{hyperxmp}
\hypersetup{
  colorlinks=true,
  linkcolor=blue,
  citecolor=blue,
  urlcolor=blue
}

\usepackage{graphicx}   
\usepackage{booktabs}   
\usepackage{amsmath}    
\usepackage{multirow}
\usepackage{listings}
\usepackage{xcolor}

\AtBeginDocument{%
  \providecommand\BibTeX{{%
    \normalfont B\kern-0.5em{\scshape i\kern-0.25em b}\kern-0.8em\TeX}}}

\setcopyright{acmcopyright}
\copyrightyear{2025}
\acmYear{2025}
\acmDOI{XXXXXXX.XXXXXXX}

\begin{document}

\title{NPHardEval4V: Dynamic Evaluation of Large Vision-Language Models with Effects of Vision}

\author{Xiang Li}
\authornote{Xiang Li and Wenyue Hua contributed equally to this research.} 
\email{lixiang0814@mail.sdu.edu.cn}
\orcid{0000-0003-1529-7057}
\affiliation{%
  \institution{Shandong University, School of Control Science and Engineering}
  \streetaddress{17923 Jingshi Road}
  \city{Jinan}
  \state{Shandong}
  \country{China}
  \postcode{250061}
}

\author{Wenyue Hua}
\authornotemark[1]
\email{wenyue.hua@rutgers.edu}
\orcid{0009-0008-2043-2704}
\affiliation{%
  \institution{Rutgers University, Department of Computer Science}
  \streetaddress{110 Frelinghuysen Road Piscataway}
  \city{New Brunswick}
  \state{NJ}
  \country{USA}
  \postcode{08854}
}

\author{Kaijie Zhu}
\email{kaijiezhu@ucsb.edu}
\orcid{0009-0002-6220-1476}
\affiliation{%
  \institution{University of California, Santa Barbara}
  \streetaddress{1210 Cheadle Hall}
  \city{Santa Barbara}
  \state{California}
  \country{USA}
  \postcode{93106}
}

\author{Lingyao Li}
\email{lingyaol@umich.edu}
\orcid{0000-0001-5888-8311}
\affiliation{%
  \institution{University of Michigan, School of Information}
  \streetaddress{500 S State St}
  \city{Ann Arbor}
  \state{Michigan}
  \country{USA}
  \postcode{48109}
}

\author{Haoyang Ling}
\email{hyfrankl@umich.edu}
\affiliation{%
  \institution{University of Michigan, School of Information}
  \streetaddress{500 S State St}
  \city{Ann Arbor}
  \state{Michigan}
  \country{USA}
  \postcode{48109}
}

\author{Jinkui Chi}
\email{chijk@umich.edu}
\affiliation{%
  \institution{University of Michigan, School of Information}
  \streetaddress{500 S State St}
  \city{Ann Arbor}
  \state{Michigan}
  \country{USA}
  \postcode{48109}
}

\author{Qi Dou}
\email{qidou@cuhk.edu.hk}
\orcid{0000-0002-3416-9950}
\affiliation{%
  \institution{The Chinese University of Hong Kong, Department of Computer Science and Engineering}
  \streetaddress{Ho Sin-Hang Engineering Building}
  \city{Sha Tin}
  \state{New Territories}
  \country{Hong Kong SAR, China}
  \postcode{999077}
}

\author{Jindong Wang}
\email{jindongwang@outlook.com}
\orcid{0000-0002-4833-0880}
\affiliation{%
  \institution{The College of William \& Mary, School of Computing, Data Sciences \& Physics}
  \streetaddress{200 Stadium Dr}
  \city{Williamsburg}
  \state{Virginia}
  \country{USA}
  \postcode{23185}
}

\author{Yongfeng Zhang}
\email{yongfeng.zhang@rutgers.edu}
\orcid{0000-0003-2633-8555}
\affiliation{%
  \institution{Rutgers University, Department of Computer Science}
  \streetaddress{110 Frelinghuysen Road Piscataway}
  \city{New Brunswick}
  \state{NJ}
  \country{USA}
  \postcode{08854}
}

\author{Xin Ma}
\authornotemark[2]
\email{maxin@sdu.edu.cn}
\orcid{0000-0003-4402-1957}
\affiliation{%
  \institution{Shandong University, School of Control Science and Engineering}
  \streetaddress{17923 Jingshi Road}
  \city{Jinan}
  \state{Shandong}
  \country{China}
  \postcode{250061}
}

\author{Lizhou Fan}
\email{leofan@cuhk.edu.hk}
\authornote{Lizhou Fan and Xin Ma are the corresponding authors.}
\orcid{0000-0002-7962-9113}
\affiliation{%
  \institution{The Chinese University of Hong Kong, Department of Psychiatry}
  \streetaddress{9 Chuen On Rd}
  \city{Tai Po}
  \state{New Territories}
  \country{Hong Kong SAR, China}
  \postcode{999077}
}








\renewcommand{\shortauthors}{Li and Hua, et al.}

\begin{abstract}

Large Vision-Language Models (LVLMs) have demonstrated impressive capabilities in multimodal understanding, yet their reasoning abilities remain underexplored. Existing benchmarks tend to focus on perception or text-based comprehension, offering limited insight into how well these models perform on structured, logic-driven tasks that require both visual and linguistic reasoning.
To address this gap, we introduce NPHardEval4V, a multimodal benchmark suite grounded in four classical NP-hard problems: Knapsack, Set Cover, Traveling Salesperson, and Vertex Cover. Each task is presented through a combination of structured visual layouts and textual prompts, designed to assess the ability of LVLMs to perform combinatorial reasoning under visual-linguistic constraints.
We evaluate a set of advanced open-source and closed-source vision-language models under a unified prompting and problem representation framework. This enables fair comparison across models and task types, while isolating key variables affecting performance.
Our results show that while these models perform reasonably well on perception-based inputs, they struggle with global optimization, abstraction, and constraint satisfaction. No single model demonstrates consistent reasoning capability across all problem types, and common failure patterns reveal fundamental limitations in current architectures.
By leveraging the structure and complexity of NP-hard problems, NPHardEval4V provides a scalable, interpretable, and challenging testbed for diagnosing reasoning behaviors in LVLMs. We hope this benchmark can support the community in building more robust, inference-capable multimodal systems. The benchmark dataset and code are available at \url{https://github.com/lizhouf/NPHardEval4V}.

\end{abstract}







\maketitle


\section{Introduction}

The evolution of Large Vision-Language Models (LVLMs) represents a significant milestone in the pursuit of artificial general intelligence (AGI), closely following advancements in Large Language Models (LLMs) \cite{fan2023bibliometric, zhao2023survey}. LVLMs introduce advanced capabilities for understanding and generating integrated textual and visual content, substantially enhancing multimedia interaction systems and cross-modal decision-making applications \cite{yin2023survey, wang2024exploring}. Reasoning, a fundamental capability required to interpret complex relationships across modalities, draw logical conclusions, and make informed decisions, is particularly critical for these models.

Evaluating the reasoning abilities of LVLMs is essential yet presents notable challenges. Reasoning tasks demand models to discern intricate relationships between visual and textual information. For instance, models may be tasked with analyzing visual differences between strings to determine transformation operations or interpreting maps to select optimal routes. Numerous existing benchmarks have evaluated LVLMs across various tasks, such as visual question answering \cite{antol2015vqa, goyal2017making, marino2019ok, hudson2019gqa}, Optical Character Recognition \cite{liu2023hidden}, robustness \cite{zhao2024evaluating}, hallucination detection \cite{liu2023hallusionbench}, and general performance evaluation \cite{xu2023lvlm, li2023seed, liu2023mmbench, yue2023mmmu}. However, these benchmarks have not specifically targeted pure reasoning capabilities, leaving a crucial evaluation gap. Existing tests often lack the rigorous, multi-step logical structure required to truly probe the limits of a model's reasoning.  Additionally, static benchmarks are vulnerable to overfitting, limiting their ability to comprehensively evaluate the evolving capabilities of the models \cite{chang2023survey}.

This paper introduces \textbf{NPHardEval4V}, a dynamic benchmark explicitly designed to rigorously evaluate the reasoning capabilities of LVLMs across diverse tasks. NP-hard problems, due to their rigorous mathematical structure, complex constraints, and often non-unique solutions, are presented as naturally suitable and stringent evaluation criteria for visual-language model reasoning. Their inherent difficulty and combinatorial nature make them superior to existing multimodal VQA tests for assessing deep, logical inference. Crucially, these problems are highly representative of real-world applications that require complex optimization and decision-making, such as logistics (Traveling Salesperson Problem), resource allocation (Knapsack Problem), and scheduling (Graph Coloring). Grounding the benchmark in these tasks allows for a more effective measurement of the practical reasoning capabilities of LVLMs. The benchmark’s dynamic nature further ensures a robust evaluation by mitigating overfitting risks.

The proposed benchmark is developed based on the NPHardEval benchmark, as presented in \cite{fan2023nphardeval}. The NPHardEval benchmark comprises nine types of algorithmic problems, categorized into three polynomial time problems, three NP-complete problems, and three NP-hard problems. Each algorithmic problem consists of 100 instances with varying difficulty levels. To enable a direct comparison between LVLMs and LLMs, we have retained the problems from NPHardEval and converted their textual descriptions into visual representations. 

We evaluate a set of state-of-the-art LVLMs, including both open- and closed-source models, across multiple reasoning tasks derived from these algorithmic problems. Our evaluation framework incorporates a structured analysis of model performance using metrics such as Recognition Accuracy (RA), Instruction-following Effective Rate (ER), and Aggregated Accuracy (AA). Furthermore, we analyze how reasoning ability varies across different complexity classes and problem difficulty levels, offering a comprehensive view into the capabilities and limitations of existing LVLMs.

Our analysis also reveals consistent patterns in model performance based on input modality. While some models benefit from concise visual prompts, others demonstrate improved reasoning when provided with richer textual context. Notably, we observe that a small subset of models, such as Gemini, exhibit robust performance across multiple input configurations, indicating advanced multimodal integration capabilities. These findings provide important insights into prompt design and model alignment for future research.

The main contributions of this work are as follows:
\begin{enumerate}
    \item \textbf{Reasoning Performance Evaluation of LVLMs:} 
    We systematically evaluate the reasoning performance of various Large Vision-Language Models (LVLMs), isolating reasoning ability from confounding factors such as image recognition and instruction-following. This analysis reveals the relative strengths and weaknesses across different models. We provide the model details of the LVLMs we evaluate in Table \ref{tab:models}.
    \item \textbf{Effects of Vision Input on LVLMs' Performance:} 
    We conduct a comprehensive ablation study on the effect of input modality, comparing three prompt types: (1) visual figures with limited textual instruction (figure+limited\_text), (2) text-only descriptions (full\_text\_only), and (3) vision-rich textual combinations (figure+full\_text). This evaluation highlights the impact of visual input on LVLMs’ reasoning performance. 
\end{enumerate}

The rest of this paper is organized as follows. Section 2 surveys related work in multimodal reasoning evaluation and benchmark design. Section 3 introduces the construction and multimodal transformation of the proposed benchmark, NPHardEval4V. In Section 4, we describe the experimental settings, including model selection, input types, and evaluation metrics. Section 5 presents experimental results and comparative analyses, followed by an examination of the influence of different input formats in Section 6. Finally, Section 7 discusses key findings, limitations, and future research directions.

\section{Related Work}

Evaluating multimodal reasoning needs benchmarks beyond perception or association. Although LVLMs perform well on captioning and VQA, they often rely on shallow heuristics. Current benchmarks rarely test structured, symbolic, or constraint-based reasoning that reflects real cognition. This section reviews related benchmarks and LVLM evaluations on complex tasks.

\subsection{Large Vision-Language Models (LVLMs) and Their Reasoning Abilities}

LVLMs \cite{wang2024exploring} can process and interpret various multimodal data streams, including imagery and textual content \cite{wu2023multimodal}. As such, LVLMs are able to surpass singularly moded LLMs and unlock new avenues for performing real-world applications. LVLMs are also more intelligent, user-centric, and holistic as compared to their LLM counterparts \cite{yin2023survey}. LVLMs can mimic the way of how humans perceive the environment by assimilating multi-sensory inputs that can complement each other. They can also foster intuitive user interactions and communications. The scope of tasks that LVLMs can assist with is significantly broader compared to LLMs, reinforcing their versatility in applications \cite{yin2023survey} such as engineering \cite{cui2024survey} and healthcare \cite{pal2024gemini}. 

Reasoning is one of the fundamental intelligent behaviors, essential for solving complex real-world tasks \cite{yu2023nature, huang2022towards}. However, even in text-centric settings, LLMs lack proper reasoning abilities, such as dealing with NP-hard (nondeterministic polynomial time) mathematical problems \cite{fan2023nphardeval}. The quest to explore and improve reasoning capabilities of Strong AI particularly within LVLMs remains a persistent challenge and a pursuit of ongoing research \cite{chen2023llava, liu2023controlllm, gong2023mindagent}.

In the domain of LVLMs, researchers have explored a range of techniques, such as instruction-tuning and prompt engineering, to enhance multimodal reasoning \cite{wang2024exploring}. The practice of instruction tuning, vital for in-context learning (ICL), has emerged as one of the key techniques in enhancing the reasoning abilities of these models. For example, frozen LLM is an initial MLLM to showcase ICL ability \cite{tsimpoukelli2021multimodal}. Later, the Flamingo model demonstrated strong ICL capabilities with a more sophisticated LLM coupled with massive-scale image and text data for its pre-training phase \cite{alayrac2022flamingo}. Prompt engineering, on the other hand, has illustrated its effectiveness in improving the reasoning abilities of LVLMs \cite{koh2023grounding, guo2023images}. As summarized by \cite{wang2024exploring}, this approach entails a variety of strategic implementations, such as representation learning \cite{koh2023grounding}, exemplar generation \cite{guo2023images}, and model interactions \cite{wang2022language}. 

\subsection{Benchmarks of Large Vision-Language Models (LVLMs)}

As the reasoning ability of LVLMs continues to advance, benchmarks have become instrumental in evaluating their performance and identifying areas that require improvement. A robust multimodal reasoning benchmark must fulfill three key criteria: (1) the integration of multimodal information, (2) the categorization of reasoning, and (3) in-depth annotations of the reasoning steps \cite{wang2024exploring}. 

Previous research has established numerous benchmarks to gauge the performance of LVLMs \cite{fu2023mme, fu2023challenger, li2023seed, liu2023hallusionbench, cao2023towards}. Among these, \cite{fu2023mme} introduced an extensive MLLM evaluation benchmark to assess models' perception and cognition skills across 14 distinct subtasks, including commonsense reasoning and code reasoning. Later, \cite{li2023seed} presented a benchmark called SEED-Bench, comprising 19,000 multiple-choice questions with precise human annotations. SEED-Bench evaluates LVLMs across 12 dimensions, capturing the ability to understand image and video modalities. In a more recent study, \cite{zhang2024benchmarking} further examined the self-consistency of MLLM responses in the presence of common corruptions. They created MMCBench, an extensive benchmark encompassing over 100 prominent LLMs. This benchmark assesses cross-modal interactions among text, images, and speech, incorporating four key generative tasks: text-to-image, image-to-text, text-to-speech, and speech-to-text.

Although existing benchmarks have evaluated the abilities of LVLMs across various dimensions, they do not provide a clear picture of the pure reasoning ability of LVLMs because factors such as recognition, knowledge amount, instruction following, and others are all combined in the presented performances in these benchmarks. Given that LVLMs' general performance are intricately dependent on their recognition process and instruction-following ability, our study aims to factor out other factors in order to see the pure reasoning process and assess them of LVLMs. Moreover, these benchmarks lack dynamic updating mechanisms, which increases the risk of LVLMs becoming overfitted to these benchmarks and restricts their ability to accurately reflect the full range of LVLMs' capabilities \cite{chang2023survey}. To address this research gap, our study proposes a dynamic assessment framework for evaluating the reasoning performance of LVLMs. In addition, we utilize the computational complexity hierarchy to rigorously assess the extent to which these LVLMs can achieve in reasoning tasks \cite{fan2023nphardeval}.

\section{NPHardEval4V: Benchmark Design}
NPHardEval4V Benchmark is built on the NPHardEval Benchmark \cite{fan2023nphardeval}, using the same set of problems while transforming the input type from textual to visual. This section first introduced the original NPHardEval Benchmark, including its question hierarchy and rotustness, and then outlines the transformation of the NPHardEval benchmark to suit the evaluation of LVLMs within NPHardEval4V, emphasizing the multimodal aspects of the tasks and the dynamic nature of the challenges. 

\subsection{NPHardEval Benchmark}
The NPHardEval Benchmark segments reasoning task into three primary computational complexity classes: P (polynomial time), NP-complete (nondeterministic polynomial-time complete), and NP-hard. These classifications serve to delineate the intrinsic difficulty and the computational resources required for solving tasks within each class, showcasing an increasing order of complexity. 

To provide a comprehensive overview of task complexity and difficulty within these classes, we introduce a hierarchical categorization illustrated in Table \ref{tab:questions}, where tasks are divided into 10 levels of progressive difficulty. This granular difficulty grading enables a nuanced assessment of model performance across a spectrum of computational challenges, thereby offering valuable insights into the capabilities and limitations of current models' reasoning abilities. 

\begin{table*}[ht]
    \caption{Complexity classes and tasks}
    \centering
    \begin{tabular}{cc}
    \toprule
        \textbf{Complexity Class} & \textbf{Task} \\
    \midrule
        \multirow{3}{*}{NP Hard (Most Complex)} & Graph Coloring Problem Optimization Version (GCP)\\
        & Traveling Salesman Problem Optimization Version (TSP)\\
        & Meeting Scheduling Problem (MSP)\\
    \midrule
        \multirow{3}{*}{NP Complete} & Knapsack Problem (KSP)\\
        & Traveling Salesman Problem Decision Version (TSP-D)\\
        & Graph Coloring Problem Decision Version (GCP-D)\\
    \midrule
        \multirow{3}{*}{P (Least Complex)} & Shortest Path Problem (SPP)\\
        & Edit Distance Problem (EDP)\\
        & Binary Search Problem (BSP)\\
    \toprule
    \end{tabular}
    
    \label{tab:questions}
\end{table*}

For each level, we then randomly generate 10 questions and this results in a total of 900 questions for one version in the dynamic benchmark. The benchmark is designed to be comprehensive and is supported by a dynamic update mechanism that ensures continuous renewal when necessary. While an infinite number of reasoning questions would ideally be required to precisely evaluate a model's capability on a specific task, practical constraints necessitate a finite sample. Based on sample size estimations \citep{maccallum1999sample, morse1992qualitative} and margin of error calculations \citep{israel1992determining, yamane1973statistics}, 100 questions per task are necessary to achieve a 95\% confidence level with a 10\% margin of error. 

\subsection{Transformation to NPHardEval4V}

To enable direct comparison with the text-only prompt input in the NPHardEval benchmark, we transform the text-only questions and represent the data part with figures. We further illustrate these transformations, with the data of the problem presented both textually and visually.

\subsubsection{Graph Data Transformation}
\begin{figure}[!h] 
    \centering
    \includegraphics[width = 0.3\textwidth]{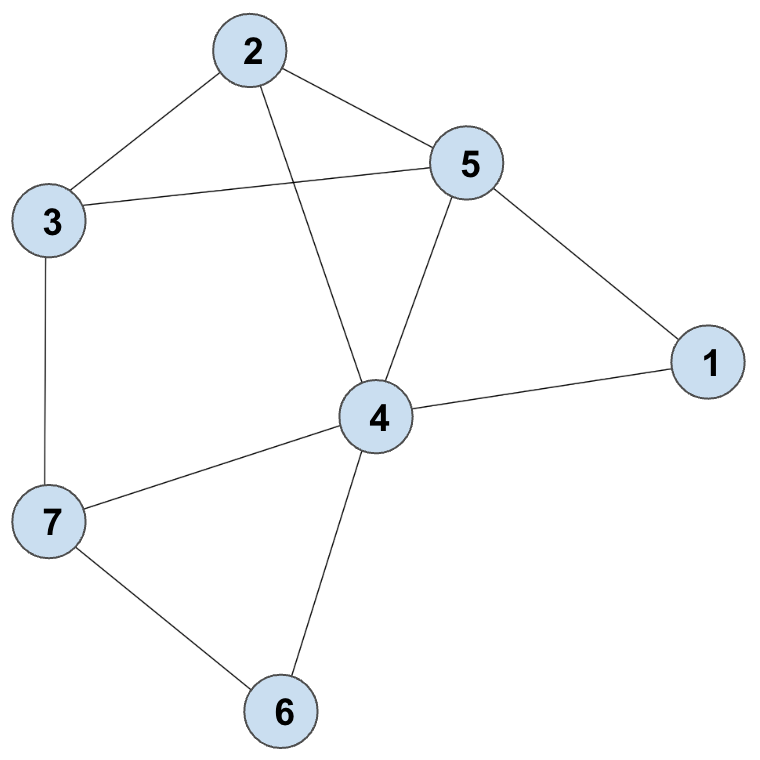}
    \caption{GCP Example}
    \label{Fig:example_gcp}
\end{figure}

The general construction of graph data problems involves providing LVLMs with both a textual description and a visual representation. For example, in the Graph Coloring Problem (see Figure \ref{Fig:example_gcp}), LVLMs are given a textual prompt alongside a figure depicting a graph, challenging them to reason across modalities. The figure is generated using Python based on the textual description of the graph coloring problem (GCP): it includes information about the nodes and the connections between them.

\begin{lstlisting}[language=HTML]
Graph coloring refers to the problem of coloring vertices of a graph in such a way that no two adjacent vertices have the same color.
There are 7 vertices 1 to 7 in a graph. You may use 3 colors with alphabets from A, B, C,... to color the graph. The graph is provided.
\end{lstlisting}

Figure \ref{Fig:example_gcp} is transformed from the textual description of the below:

\begin{lstlisting}[language=HTML]
Vertex 1 is connect to 5, 4.
Vertex 2 is connect to 3, 4, 5.
Vertex 3 is connect to 5, 7.
Vertex 4 is connect to 1, 2, 6, 7.
Vertex 5 is connect to 1, 3, 4.
Vertex 6 is connect to 4, 7.
Vertex 7 is connect to 3, 4, 6.
\end{lstlisting}

Figures in algorithmic questions such as SPP, GCP$\_$D, TSP, and TSP$\_$D are transformed from text similarly: these figures include graphs containing nodes and edges; in the case of TSP and TSP$\_$D, the weights on the edges are also included.

\subsubsection{Linear Data Transformation}

\begin{figure}[!htbp] 
    \centering
    \includegraphics[width = 0.35\textwidth]{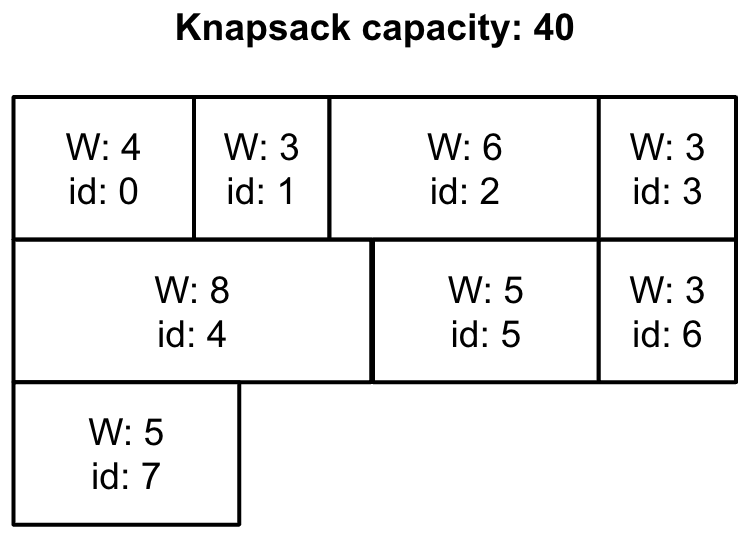}
    \caption{KSP Example}
    \label{Fig:example_ksp}
\end{figure}

Similarly, linear data problems like the Knapsack Problem (see Figure \ref{Fig:example_ksp}) are presented with both a prompt and a visual representation. In order to strictly align the input content across modalities, our generation scripts ensure that the visual representation is a neutral and information-preserving rendering of the textual data. For instance, in the Knapsack problem, we use Python to create blocks with sizes and labels that precisely correspond to the weights and IDs of the items described in the text. This programmatic approach is designed to ensure informational equivalence and avoid one modality inadvertently containing more prompt information than the other.


\begin{lstlisting}[language=HTML]
The Knapsack Problem (KSP) asks whether a subset of items, each with a given weight and value, can be chosen to fit into a knapsack of fixed capacity, maximizing the total value without exceeding the capacity. Determine if a subset of items can be selected to fit into a knapsack with a capacity of 40, maximizing weight without exceeding the capacity. Item weights are presented by the top number after 'W:'. Item IDs are shown as numbers below the weights after 'id:'.
\end{lstlisting}

Figure \ref{Fig:example_ksp} is transformed from the textual description of the below:

\begin{lstlisting}[language=HTML]
Item 0 has weight 4.
Item 1 has weight 3.
Item 2 has weight 6.
Item 3 has weight 3.
Item 4 has weight 8.
Item 5 has weight 5.
Item 6 has weight 3.
Item 7 has weight 5.
\end{lstlisting}

In the MSP problems, we generate figures that visually represent the availability of each person in a calendar format corresponding to the textual description of the problem. Specifically, we use Python to create a calendar view that displays the availability of each person based on the input data. 

For algorithmic problems such as EDP and BSP, we do not employ a specialized visualization method. Instead, we simply display the relevant string of characters or string of numbers in the image.

In summary, we employ various visualization functions to transform textual descriptions into visual representations for different algorithmic problems. In the prompt, we provide a combination of textual and visual information to the model, including an instructional prompt and a general introduction of the algorithm, along with an image that represents the specific problem to be solved. This approach is demonstrated in the previous examples for GCP and KSP. By providing both textual and visual information, we aim to evaluate the reasoning abilities of LVLMs in handling complex problems that require both visual and textual understanding.

\subsection{Dynamic Benchmark Mechanism}
A core feature of NPHardEval4V is its dynamism, designed to prevent benchmark overfitting and provide a continuous challenge for evolving models. This dynamism is  implemented through a programmatic generation pipeline.

\textbf{Implementation of Dynamism: }The benchmark is powered by a suite of algorithms that can generate new, unique problem instances on demand for each task (e.g., KSP, TSP). For each new version of the benchmark, our generator program is executed to create a fresh dataset of 900 questions (9 tasks × 10 difficulty levels × 10 questions per level). This ensures that models cannot simply memorize solutions from previous versions. The generation process guarantees that while the underlying logic of the task remains the same, the specific parameters (e.g., graph structures, item weights, city coordinates) are novel, ensuring the unpredictability of samples.

\textbf{Version Control and Update Frequency: }To maintain consistency and track progress over time, each generated dataset is versioned. We recommend a bi-annual or annual update frequency, which strikes a balance between keeping pace with rapid model development and providing a stable target for the research community.

\section{Experimental Setting}

To address the outlined research questions, our experimental setting systematically assesses the performance of various LVLMs in recognition and reasoning tasks and investigates the impact of vision and text inputs on LVLMs' performance. The models we test on are presented in Table \ref{tab:models}. We detail the experimental setup corresponding to each research question, describing the models evaluated, the nature of the tasks, the methodology of the performance assessment, and the structure of the benchmark.

\begin{table*}[ht]
    \centering
    \begin{tabular}{ccc}
    \toprule
        \textbf{Model} & \textbf{Number of Parameters (B)} &  \textbf{Access} \\
    \midrule
        GPT-4o \cite{openai_gpt4o_systemcard} & Unknown & Close source \\
        Gemini-2.0-flash-exp \cite{GeminiPro} & Unknown & Close source \\
        claude-3.7-sonnet \cite{claude3_7_sonnet} & Unknown & Close source \\
        Llama-3.2-vision \cite{huang2024empirical} & 90 & Open source \\
        InternVL3 \cite{zhu2025internvl3} & 78 & Open source \\
        Llava-qwen2 \cite{li2024llavaonevisioneasyvisualtask} & 72 & Open source \\
        Qwen2.5-VL \cite{qwen2.5-VL} & 72 & Open source \\
        Qwen2-VL \cite{Qwen2VL} & 72 & Open source \\
        Nvlm-D \cite{dai2024nvlmopenfrontierclassmultimodal} & 72 & Open source \\
        Ovis2 \cite{lu2024ovisstructuralembeddingalignment} & 34 & Open source \\
        DeepSeekVL2 \cite{wu2024deepseekvl2mixtureofexpertsvisionlanguagemodels} & 27 & Open source \\
        Gemma \cite{team2025gemma} & 27 & Open source \\
        GLM-4V \cite{glm2024chatglm} & 9 & Open source \\
        Ola \cite{liu2025olapushingfrontiersomnimodal} & 7 & Open source \\
        Point1.5-qwen2.5 \cite{liu2024points15buildingvisionlanguagemodel} & 7 & Open source \\
        
    \toprule
    \end{tabular}
    \caption{LVLMs metadata}
    \label{tab:models}
\end{table*}

\subsection{Recognition and Reasoning Performance Evaluation}
In the initial experiment, the objective is to assess the performance of Large Vision-Language Models (LVLMs) in terms of recognition and reasoning across a range of benchmark problems. To this end, zero-shot prompts are utilized to evaluate the inherent recognition and reasoning capabilities of each model. The performance of the models is then analyzed and compared across the benchmark problems to draw conclusions about their relative strengths and weaknesses in recognition and reasoning tasks.

\subsubsection{Recognition Experiment}
In order to precisely assess the capacity for reasoning in Large Vision-Language Models (LVLMs), it is essential to first comprehend their proficiency in image recognition, as accurate reasoning can only be established upon correct recognition. Therefore, an evaluation of recognition capabilities is initially conducted. For each question, both a visual representation of the description and a textual representation are provided. Subsequently, the MLLM is presented with both the visual and textual representations and is asked to determine whether they correspond to the same question. To mitigate the influence of randomness, the model is prompted five times, each time with a distinct random seed. The mean and variance of the model's responses are then calculated, and it is concluded that the model demonstrates recognition capabilities if it correctly identifies the correspondence between the visual and textual representations more than half of the time, i.e., answering 'yes' three or more times. As an preprocess for analyzing reasoning ability, recognition experiment reuslts and analysis are provided in Appendix \ref{Appendix:recog}

\subsubsection{Reasoning Experiment 1, the Default Setup (Vision with Instructional Text)}
This experiment evaluates the overall performance of MLLM on a benchmark consisting of various tasks. In the setting, a textual prompt is provided that includes a general introduction to the question and the format in which the answer should be given. Subsequently, an image is presented that pertains to the specific question being asked. The models are then tasked with processing both the textual and visual information in order to generate accurate responses. The performance of the LVLMs is evaluated based on their ability to correctly answer the questions, taking into account both their recognition and reasoning capabilities. The results of this experiment provide insights into the overall effectiveness of the models in handling multimodal tasks and their potential applications in real-world scenarios.

Below is an example for an SPP problem:

\begin{lstlisting}[language=HTML]
The Shortest Path Problem (SPP) involves finding the shortest path between two nodes in a weighted graph.
You need to find the shortest path between node 0 and node 3 in a graph. The graph's edges and their weights are given.
Please provide the shortest path from 0 to 3 and its total distance. Offer a concise step-by-step explanation of your reasoning process. Aim for brevity and clarity in your response.
Your output should be enclosed within <root></root> tags. Include your reasoning in <reasoning></reasoning> tags and the final path and total distance in <final_answer></final_answer> tags, like <final_answer>{'Path': 'START->...->END', 'TotalDistance': 'INT_TOTAL_DISTANCE'}</final_answer>.
\end{lstlisting}

\begin{figure}[ht]
    \centering
    \includegraphics[scale=0.1]{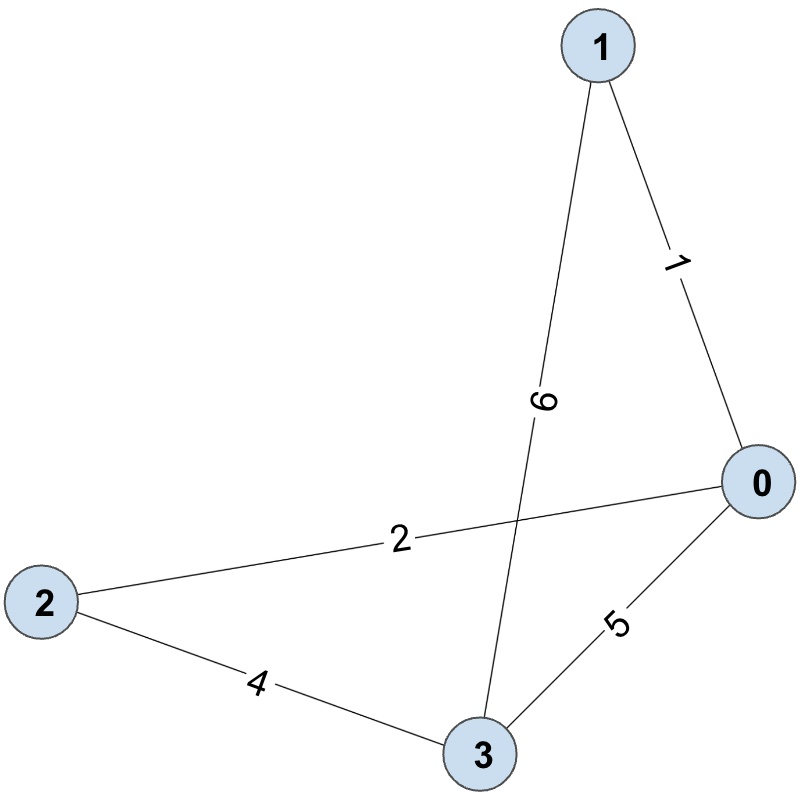}
    \caption{Image of an Short Path Problem example}
    \label{fig:example_1}
\end{figure}

In order to accurately evaluate the reasoning performance, it is necessary to isolate the specific aspect of reasoning ability from other factors that may influence the overall performance: we employ a filtering process that removes datapoints where the model fails to recognize the given input or fails to provide a parsable result, which indicates a failure in instruction-following. After filtering out these two factors, we can obtain a more accurate picture of the LVLMs' pure reasoning ability.

\subsection{Ablation on Multimodal prompts}
The goal of this experiment is to investigate the impact of prompt modality on the reasoning performance of LVLMs. Specifically, we aim to assess whether the inclusion of visual or textual inputs can enhance the models' problem-solving abilities.
To achieve this, we conduct an ablation study, varying the modality of the prompts provided to the LVLMs by pure textual prompt and textual + visual prompt. By comparing the models' performance across different prompt modalities, we aim to identify which modality leads to the best performance and evaluate the contribution of each modality to the overall performance.

\subsubsection{Reasoning Experiment 2, the Text-only Setup (Instructional and Data Text)}
In order to assess whether visual representations can be helpful, we first need to see the general problem-solving performance using pure textual prompt. In this experiment, we give LVLMs pure textual description of the problem containing the description of a specific question. 

In the SPP example, the pure textual prompt is shown as below:

\begin{lstlisting}[language=HTML]
The Shortest Path Problem (SPP) involves finding the shortest path between two nodes in a weighted graph.
You need to find the shortest path between node 0 and node 3 in a graph. The graph's edges and their weights are given.
Please provide the shortest path from 0 to 3 and its total distance. Offer a concise step-by-step explanation of your reasoning process. Aim for brevity and clarity in your response.
Your output should be enclosed within 
<root></root> tags. Include your reasoning 
in <reasoning></reasoning> tags and the final 
path and total distance in <final_answer>
</final_answer> tags, like <final_answer>
{'Path': 'START->...->END', 'TotalDistance': 
'INT_TOTAL_DISTANCE'}</final_answer>.

In this graph:
The distance between node 0 and 1 is 1,
the distance between node 1 and 3 is 9, 
the distance between node 0 and 3 is 5,
the distance between node 0 and 2 is 2,
the distance between node 2 and 3 is 4.
\end{lstlisting}

\subsubsection{Reasoning Experiment 3, the Vision-rich-text Setup (Vision with Instructional and Data Text)}
Visual aids can be helpful in solving complex problems for humans. For instance, when dealing with a GCP, sketching a diagram of the graph and attempting different coloring schemes can aid in determining whether a valid solution exists. In this experiment, we aim to investigate whether providing MLLM with both textual and visual representations of a problem can improve their performance in solving it.
To do so, we present the LVLMs with a complete textual description of the problem, followed by an accompanying image that depicts the problem visually. Our hypothesis is that the image can serve as a supplementary source of information to the textual description, potentially enhancing the model's understanding of the problem and enabling it to generate more accurate solutions. By incorporating both textual and visual inputs, we aim to evaluate the LVLMs' ability to integrate and process multimodal information in complementary with each other, which is a crucial aspect of their overall performance.

\subsection{Evaluation Metrics}
\label{sec:eval}

\begin{figure*}[h] 
    \centering
    \includegraphics[width = 1\textwidth]{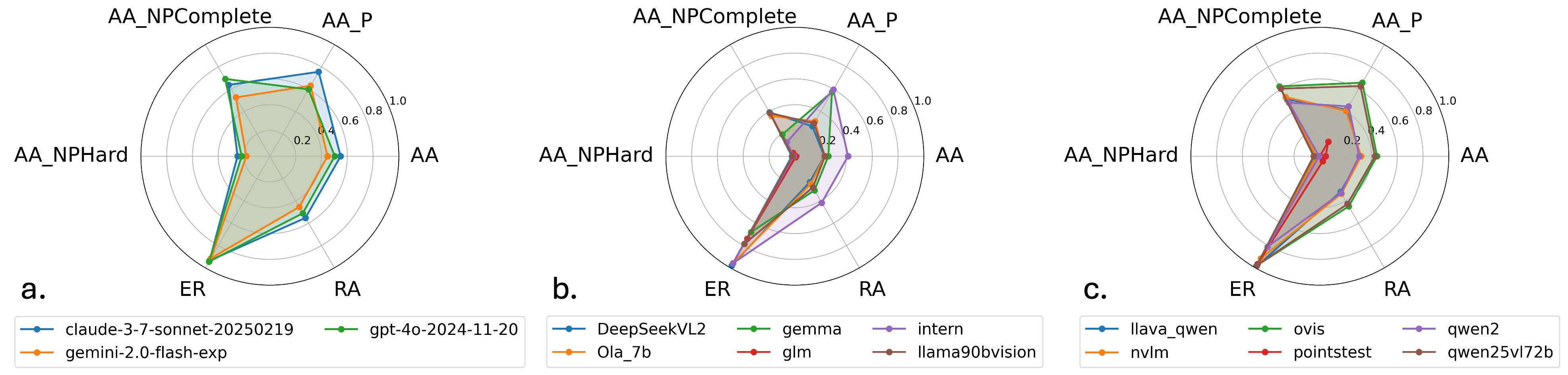}
    \vspace{-10pt}
    \caption{Multimodal Large Language Models's performance on recognition (RA), Instruction-following (ER), and reasoning (AA) on polynomial time (P), NP-complete, and NP-hard problems. All scores range from 0 to 1, with higher values indicating better performance. The a plot shows closed-source models (Claude-3.7, Gemini-2, GPT-4o), while the b and c plots show two groups of open-source models. Larger radar areas indicate stronger overall capabilities, while smaller or skewed shapes reveal specific weaknesses, particularly in handling harder reasoning tasks.}
    \vspace{-10pt}
    \label{Fig:spider}
\end{figure*}

\begin{figure}[ht] 
    \centering
    \includegraphics[width = 0.5\textwidth]{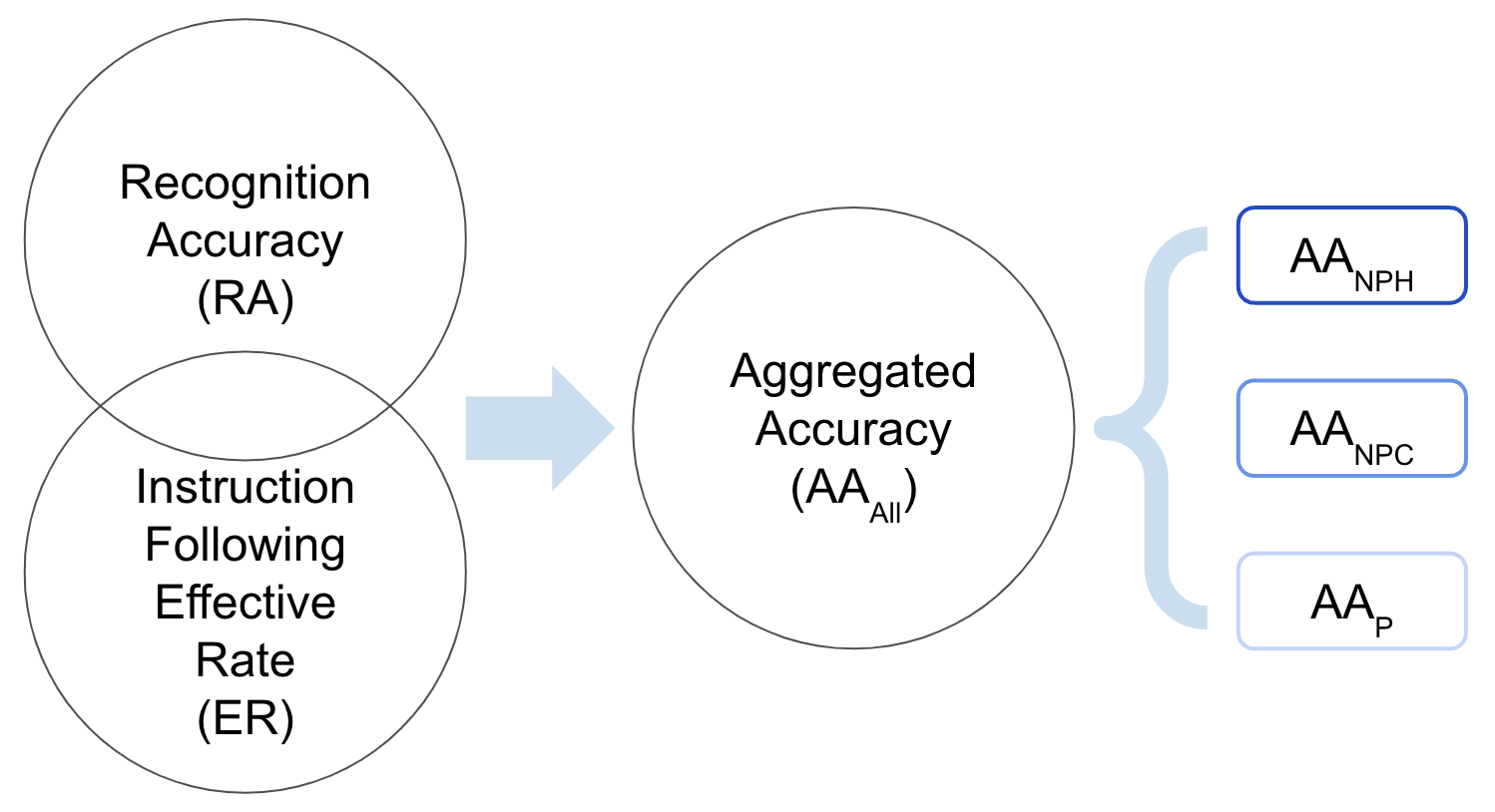}
    \vspace{-10pt}
    \caption{Metrics Sequence}
    \vspace{-10pt}
    \label{Fig:metrics}
\end{figure}

Our evaluation methodology incorporates a suite of metrics to assess the reasoning ability of LVLMs. Building upon the metrics established by \cite{fan2023nphardeval}, we expand our analysis to include a novel metric specifically tailored to quantify and rule out noise in analyzing models' reasoning abilities. This sequence of metrics will start with the Recognition Accuracy (RA) metric, evaluating the recognition accuracy of the problems in the visual prompts, i.e., if the LVLMs can recognize what are in the questions. Among those questions that a MLLM can accurately recognize the input, we then use the Instruction-following Effective Rate (ER) metric, calculating the average chances that a MLLM cannot yield an output compatible with the rule-based answer parser. Finally, the Aggregated Accuracy (AA) will calculate the weighted accuracy of LVLMs' answers correctness among those recognizable and parsable scenarios, aggregated with RA and ER. Figure \ref{Fig:spider} and Figure \ref{Fig:metrics} shows the sequence and relationship of these metrics.

\subsubsection{Recognition Accuracy (RA)}
Recognition Accuracy evaluates the LVLMs' capability to accurately interpret the visual information presented in the prompts. This metric is crucial as it forms the foundation for any subsequent reasoning or decision-making processes the model undertakes. It is quantified by the ratio of prompts that are correctly recognized by the model to the total number of prompts presented, and is defined as:
$$ RA = \frac{\sum_{i=1}^{N} C_i}{N} $$
where \( C_i \) is a binary indicator of correct recognition (1 if the \( i^{th} \) prompt is correctly recognized, 0 otherwise), and \( N \) is the total number of prompts.

\begin{figure*}
    \centering
    \includegraphics[width = 1\textwidth]{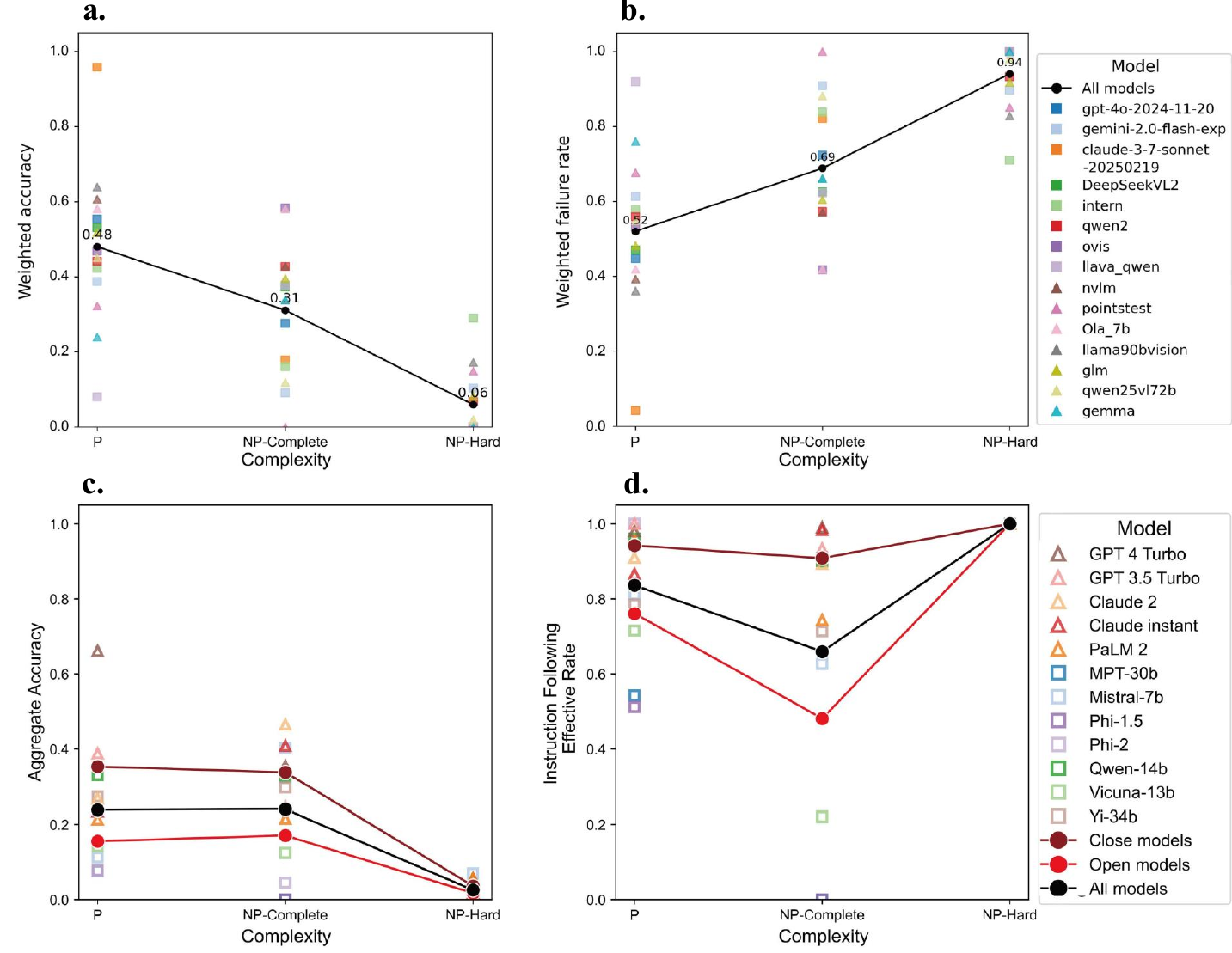}
    \vspace{-10pt}
    \caption{MLLM: a. Reasoning abilities performance excluding the effects of recognition and instruction following on figure+limited$\_$text representation of questions. b. Instruction-following effective rate. LLM: c. Reasoning abilities performance excluding the effect of instruction following on pure textual description of questions. d. Instruction-following effective rate}
    \vspace{-10pt}
    \label{Fig:AA_ER_line}
\end{figure*}


\subsubsection{Instruction-following Effective Rate (ER)}
Instruction-following Effective Rate measures the average likelihood that an MLLM's response adhere to the expected output format, thus being compatible with a rule-based answer parser. This metric is crucial for gauging the models' reliability in producing solutions complying to standard parsing:
$$ ER = \frac{\sum_{i=1}^{N} F_i}{N} $$
where \( F_i \) is parsing Instruction-following status for the \( i^{th} \) problem, where \( F_i=1 \) if the parsing task successes and \( F_i=0 \) otherwise, and \( N \) is the total number of problems.

\subsubsection{Aggregated Accuracy (AA)}
Aggregated Accuracy takes into account the correct recognition of prompts and the successful parsing of responses to evaluate the correctness of LVLMs' answers. This metric is a weighted measure that integrates the Recognition Accuracy and the inverse of the Failure Rate:
$$ AA = \frac{\sum_{i=1}^{N} (w_i \times A^{'}_{i} \times RA_i \times (ER_i))}{\sum_{i=1}^{N} w_i} $$
where \( w_i \) represents the difficulty weight for the \( i^{th} \) level, \( A^{'}_{i} \) represents the accuracy in recognizable and parsable questions at \( i^{th} \) level, \( RA_i \) is the Recognition Accuracy for the \( i^{th} \) level, and \( ER_i \) is the Instruction-following Effectiveness Rate for the \( i^{th} \) level. The Aggregated Accuracy metric will be used as the main evaluation metric for the overall performance and the complexity-specific submetrics, demoted as \( AA \) or \( AA_{All} \), \( AA_{NPH} \), \( AA_{NPC} \), and \( AA_{P} \).


\begin{figure*}[!ht] 
    \centering
    \includegraphics[width = 1\textwidth]{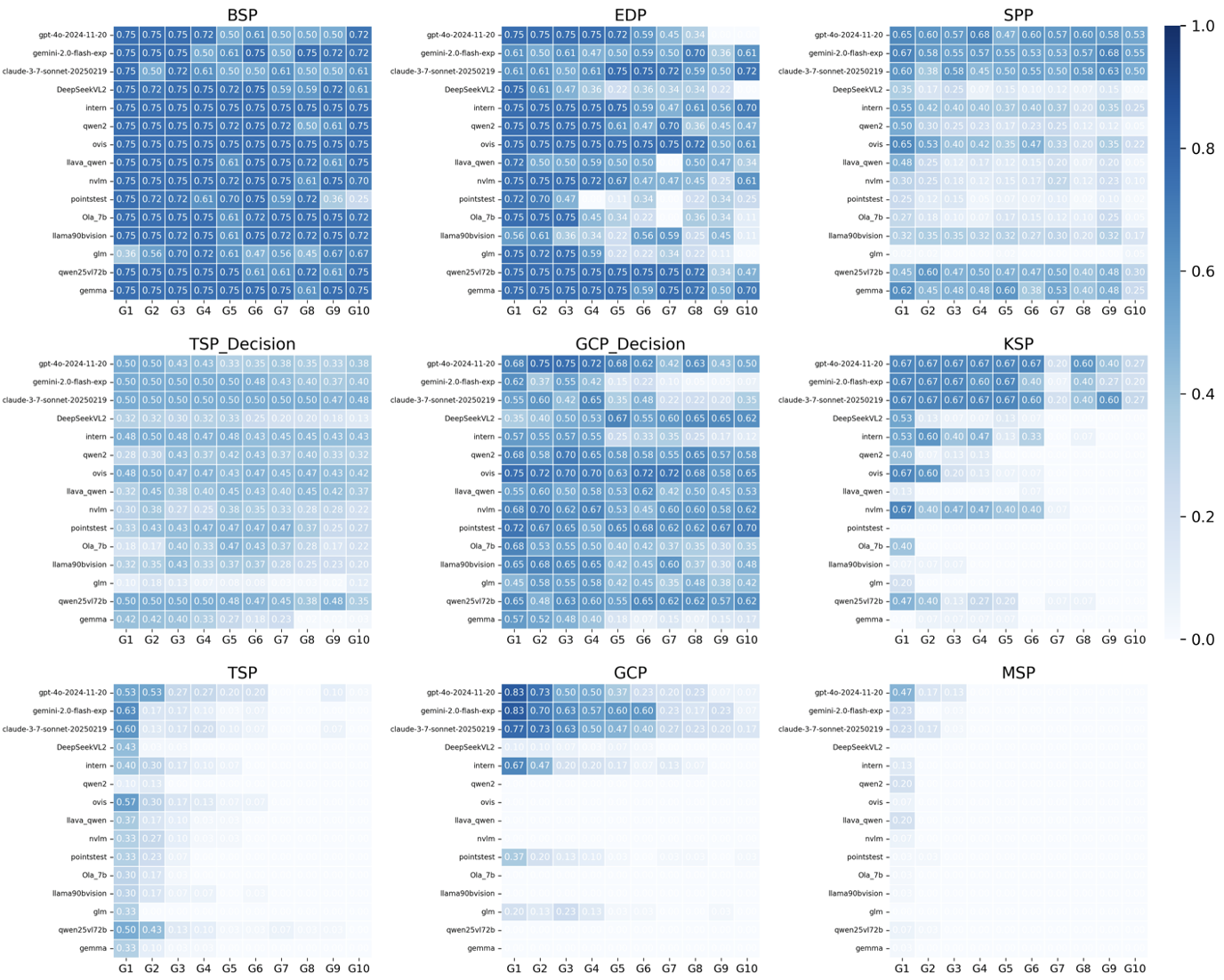}
    \vspace{-10pt}
    \caption{Reasoning abilities across models, complexity levels, and difficulty levels}
    \vspace{-10pt}
    \label{Fig:AA_heatmap}
\end{figure*}

\section{Results}
This section presents the findings from our experiments using the NPHardEval4V benchmark. We evaluate and compare the reasoning abilities of various Large Vision-Language Models (LVLMs), considering factors such as model type (open vs. closed source), task complexity, question difficulty, and the comparison between LVLMs and traditional LLMs.

\subsection{Reasoning Abilities of LVLMs}
To better understand the reasoning potential of LVLMs, we analyze their performance across several axes, starting with the difference between open-source and closed-source models. As Figure \ref{Fig:spider} and Figure \ref{Fig:AA_ER_line} indicate, there is a huge difference in performance between close source and open source LVLMs, with close source models exhibiting superior performance in all tasks, irrespective of complexity class. Specifically, demonstrated in Figure \ref{Fig:AA_heatmap}, the Claude-3.7 model consistently outperforms its counterparts, including the other close source model Gemini and GPT-4o, across the board. A detailed inspection of the heatmap reveals that Claude-3.7 maintains a higher aggregated accuracy and a more effective Instruction-following rate than Gemini, which suggests that the proprietary nature in the Claude-3.7 model may contribute to enhanced reasoning performance. This is especially evident in the NP-hard category, where Claude-3.7 shows remarkable resilience in reasoning accuracy compared to Gemini, which exhibits a notable decline.

As Figure \ref{Fig:AA_heatmap} indicates, the reasoning capabilities of LVLMs are inversely proportional to the complexity of the tasks. On simpler P problems, these models show commendable performance. However, as the complexity escalates to NP-complete and further to NP-hard problems, a clear and expected downtrend in reasoning ability is observed. This trend is consistent across all LVLMs, signifying a universal challenge in tackling higher-order complexity with current multimodal reasoning architectures.

In addition to the overall trend, there are two notable findings. First, none of the fifteen models are capable of solving two of the NP-hard problems, TSP and MSP. Second, the Qwen2.5-vl, Ovis and Internvl model performs unexpectedly well on the TSP-D problem -- close to or even better than closed-source models. This may due to the specific training data relevant or similar to the TSP-D problem.

When focusing on individual reasoning tasks and considering models that can at least address the simplest questions, we notice a degradation in performance in correlation with increasing question difficulty. As Figure \ref{Fig:AA_heatmap} shows, models like Gemini display a high success rate on easier questions within tasks like GCP, but this success rate diminishes as the difficulty level of the questions increases. This pattern underscores the models' limitations and suggests that even the most capable LVLMs struggle to maintain their reasoning prowess as task complexity intensifies.

Figure \ref{Fig:AA_ER_line}  provide a direct comparison between the current top-performing models in LVLMs and LLMs, including both closed-source and open-source models. For LLMs, we selected the models GPT-4-turbo, GPT-3.5-turbo, Claude-2, Claude-instant, PaLM-2, MPT-30b, Mistral-7b, Phi-1.5, Phi-2, Qwen-14b, and Yi-34b \cite{fan2023nphardeval}. The results indicate that LLMs lag behind LVLMs in reasoning tasks. Specifically, the aggregated accuracy in P problems is approximately 0.4 in LLMs and 0.48 in LVLMs. However, the decrease in aggregated accuracy is more pronounced in LVLMs compared to LLMs. While LLMs maintain a relatively consistent weighted accuracy in NP-complete and P problems, LVLMs' weighted accuracy decreases 35.42\%, from approximately 0.48 to 0.31. These findings highlight the need for further research and development to improve the reasoning abilities of LVLMs.

The results of our study indicate that the integration of multimodal data processing in LVLMs does not necessarily lead to improved reasoning capabilities. In fact, our findings reveal that the current development of LVLMs falls short in comparison to LLMs, with significantly weaker performance in reasoning tasks. This highlights the need for further research and development efforts aimed at enhancing the reasoning abilities of LVLMs. 

\subsection{Impact of Vision and Text Input}

\begin{figure}[ht] 
    \centering
    \includegraphics[width = 0.8\textwidth]{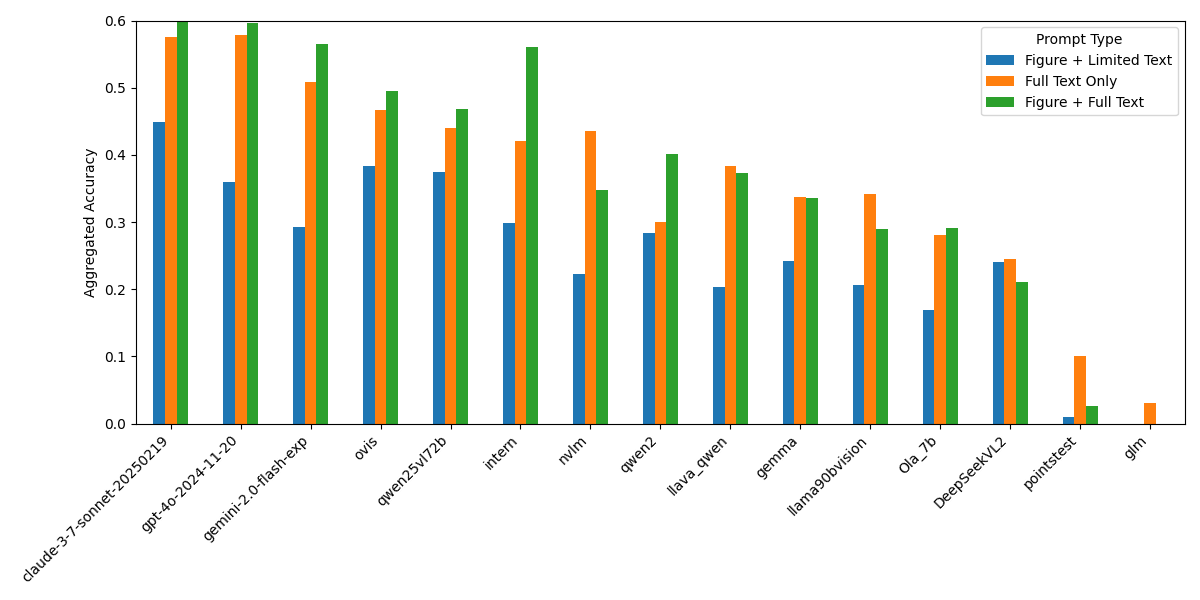}
    \vspace{-10pt}
    \caption{Reasoning abilities across prompt types}
    \vspace{-10pt}
    \label{Fig:AA_prompt_type_bar}
\end{figure}


In evaluating the effect of different combinations of visual and textual inputs on the reasoning abilities of LVLMs, we observe varied responses across the models. Figure \ref{Fig:AA_prompt_type_bar} indicates that a significant majority of the models, including both close source models like Claude-3.7 and open source models such as internvl, exhibit the highest levels of reasoning accuracy with the Default Setup, which pairs figures with limited instructional text (figure+limited\_text). This suggests that their textual understanding is relatively weaker compared to their visual understanding, as the use of images to represent the question seems to be more effective than textual representation. Furthermore, when combining figure with full text, the models' performance does not improve and instead decreases, indicating that the addition of excessive textual information may hinder their reasoning abilities. These findings highlight the importance of considering the appropriate combination of visual and textual inputs in designing tasks for LVLMs, as well as the need for further research to enhance their textual understanding abilities.

In contrast, the Claude-3.7 model stands out distinctly by demonstrating superior reasoning performance with both the Text-only Setup (full\_text\_only) and the Vision-rich-text Setup (figure+full\_text). This unique pattern indicates that Claude-3.7 may have a more advanced approach to processing and integrating comprehensive textual information, which is further enhanced by the presence of visual data. It's noteworthy that while the addition of full textual context to the visual prompts (Vision-rich-text Setup) improves Claude-3.7's performance, it does not have the same positive effect on the other models tested, potentially due to the increased task complexity and vision recognition demands.

These findings reveal that the interplay between visual and textual inputs is not uniform across LVLMs and that certain models may be specially attuned to leverage text, visuals, or a combination of both to maximize their reasoning proficiency. It underscores the importance of prompt design in MLLM performance and calls for further investigation into how different models process multimodal information.

\subsection{Failed Case Analysis}
While quantitative metrics show high failure rates on complex tasks, a qualitative analysis is crucial for understanding why models fail. We analyzed typical error responses and identified several common failure patterns: reasoning chain errors, instruction-parsing failures, and visual grounding errors.

Below, we present a sentence-by-sentence analysis of a typical erroneous response for a Knapsack Problem (KSP) instance.
\begin{lstlisting}[language=HTML]
Task: Given 9 items with weights {9, 6, 1, 1, 7, 6, 4, 9, 6} and values {8, 8, 4, 6, 7, 8, 6, 8, 1}, select a subset of items to maximize total value without exceeding a knapsack capacity of 45.
Correct Answer: Select items with weights {9, 6, 1, 1, 7, 6, 4, 9} (items 0, 1, 2, 3, 4, 5, 6, 7) for a total value of 55.
Typical Incorrect Model Response:
<root>
<reasoning>
The objective is to maximize the weight without exceeding the knapsack capacity of 45 while maximizing value. We use a dynamic programming approach to determine the optimal selection.\n\n        
Listing items:\n        
- (9,8)  -> ID 0 \n        
- (6,8)  -> ID 1 \n        
- (1,4)  -> ID 2 \n        
- (1,6)  -> ID 3 \n        
- (7,7)  -> ID 4 \n        
- (6,8)  -> ID 5 \n        
- (4,6)  -> ID 6 \n        
- (9,8)  -> ID 7 \n        
- (6,1)  -> ID 8 \n        
Sorting by efficiency (Value/Weight):\n        
- ID 3: (1,6) -> 6.00\n        
- ID 2: (1,4) -> 4.00\n        
- ID 6: (4,6) -> 1.50\n        
- ID 4: (7,7) -> 1.00\n        
- ID 0: (9,8) -> 0.89\n        
- ID 1: (6,8) -> 1.33\n        
- ID 5: (6,8) -> 1.33\n        
- ID 7: (9,8) -> 0.89\n        
- ID 8: (6,1) -> 0.17\n\n        
Using dynamic programming to maximize value:\n        
The optimal subset is {0, 1, 2, 3, 4, 5, 6} summing up to weight 45 and total value 47.
 </reasoning> <final_answer>{'SelectedItems': [0,1,2,3,4,5,6], 'TotalValue': 47}</final_answer> </root>
\end{lstlisting}

\subsubsection{Correct Intermediate Steps, Wrong Reasoning Strategy:} The model correctly identifies all items, their weights, and values from the input. It also parses the knapsack capacity and performs valid arithmetic to ensure the selected items’ total weight does not exceed the constraint (the sum is exactly 45). Furthermore, the computed total value for the selected items—47 for items {0, 1, 2, 3, 4, 5, 6}—is accurate. However, while these intermediate calculations are correct, the overall selection strategy is suboptimal. The model’s method of sorting items by value/weight (“efficiency”) and then applying a dynamic programming approach misses the globally optimal solution.
\subsubsection{Source of Failure:} Reasoning Chain Error.
The primary source of failure is in the reasoning strategy. Although the model claims to use dynamic programming, the actual logic reflects a greedy or near-greedy selection process based on item efficiency. This heuristic does not guarantee an optimal solution for the Knapsack problem. In this example, the model selects items {0, 1, 2, 3, 4, 5, 6} (total value 47), but fails to include item 7, which would increase the total value to 55 without exceeding capacity. The model does not properly explore all combinations to maximize the objective, demonstrating an incomplete implementation of dynamic programming or a misunderstanding of the required exhaustive search for optimality.
\subsubsection{Discussion of Hallucination:} This example illustrates a form of hallucination: the model outputs a reasoning trace that appears algorithmic and plausible, but actually misrepresents both its method (“using dynamic programming”) and its effectiveness. The step-by-step explanation is internally consistent but ultimately implements a shortcut that cannot always solve the problem optimally. This highlights the risk of accepting well-written explanations at face value—surface plausibility may mask underlying errors in logic or algorithm selection.

\section{Conclusion and Discussion}

In this paper, we have expanded upon the initial introduction of NPHardEval4V, a dynamic and comprehensive benchmark that scrutinizes the reasoning capabilities of Large Vision-Language Models (LVLMs) against the backdrop of computational complexity. Our aim has been to dissect and understand the multifarious abilities of LVLMs, particularly focusing on their capacity for reasoning in response to visual-textual prompts.

The experimental results, complemented by the transformed data on the impact of vision and text input, suggest that the performance of LVLMs is contingent upon not only the complexity of the task but also the nature of input they are provided. We have observed that while most models, including close source models like Claude-3.7 and open source ones like Internvl, exhibit optimal performance with the Default Setup (figure with limited instructional text), it is the Gemini model that stands out, showing a marked improvement in reasoning ability when provided with text-only and vision-rich-text (figure with full textual descriptions) prompts.

NPHardEval4V operates under the principle that benchmarks must be as dynamic as the models they assess. This is reflected in the benchmark's diverse suite of tasks, which compel LVLMs to demonstrate not just recognition but sophisticated reasoning and the ability to learn adaptively. This benchmark highlights the critical need for LVLMs to process and learn from both visual and textual data effectively, underscoring the importance of dynamic evaluation tools that can truely evaluate LVLMs' progression.

In conclusion, the insights garnered from NPHardEval4V shed light on the present competencies and constraints of LVLMs. While we have detailed the specific outcomes of our experiments, the broader trends emphasize the necessity of dynamic and stringent testing to deepen our comprehension and further the advancement of AI. As we continue to extend the possibilities of LVLMs, benchmarks like NPHardEval4V will be instrumental in steering the evolution of models that are not just powerful but also multifaceted, adaptable, and truly intelligent. This study acts as a clarion call for the AI community to recognize the complexities of multimodal reasoning and to strive for models that can seamlessly integrate diverse inputs to reason and learn more like humans do.

\subsection{Limitations}

The benchmarking process inherently favors models like claude-3.7 over others due to the varied architectural strengths, which may not be fully captured in a uniform evaluation setup. Despite our attempts to level the playing field, certain models are naturally more adept at specific types of reasoning tasks due to their difference in training data, which could skew the overall comparative analysis. For example, claude-3.7's superior performance in text-only setups may reflect an underlying bias in the benchmark towards linguistic reasoning over multimodal reasoning.


Furthermore, model selection is concentrated on mainstream large models, lacking the inclusion of models that leverage specialized visual priors, such as SAM or DETR-ViT based architectures. Evaluating such models could reveal different performance patterns. Similarly, the diversity of tasks on the benchmark is limited. The current tasks are grounded in classic algorithmic problems and do not cover other important areas like image semantic reasoning, visual common-sense reasoning, chart understanding, or multi-step procedure generation from visual guides. This may lead to an incomplete picture of models’ true reasoning capacities. 

The design of image and text inputs, despite our efforts to ensure informational equivalence, may still introduce biases. Images can simplify certain details, which makes some models 'visually easier to answer.' To properly calibrate this effect, future work should add a human baseline or a o3 + OCR baseline. This would help determine whether humans or other powerful systems also exhibit a preference for a particular modality, providing a stronger reference for assessing fairness.

Finally, evaluation metrics focus on accuracy-based indicators (RA, ER, AA). While useful, they do not provide a deep view into the model's internal processes. The analysis lacks a systematic evaluation of the interpretability of reasoning chains, which could be enhanced by visualizing the model's reasoning trace or performing attribution analysis.

The study highlights the variability in model performance based on prompt type, suggesting a significant dependence on how questions are framed. This raises concerns about the models' robustness and their ability to generalize across different input conditions. It is unclear whether the observed performance reflects true reasoning ability or an artifact of the models' sensitivity to specific prompt structures.

The weaker performance of LVLMs in comparison to LLMs suggests that the integration of multimodal data is not always be beneficial for reasoning tasks. This observation implies limitations in how current models process and integrate multimodal information, indicating a misalignment between model architecture and task requirements.

\subsection{Research Outlook}

Investigating models' learning curves over extended periods could offer valuable insights into their potential for growth and adaptation. Long-term studies that track how models evolve with additional training data or through incremental learning can provide a deeper understanding of the underlying mechanisms that drive progress in LVLMs and identify the key factors that contribute to long-term success.

To complement long-term studies of model development, future work should also broaden the taxonomy of reasoning tasks. This includes incorporating challenges such as visual common-sense reasoning, chart and diagram interpretation, and multi-step procedural generation—areas that better reflect the diverse applications of modern LVLMs.

Alongside task expansion, deeper interpretability is crucial. One direction is to introduce reasoning trace visualizations, enabling researchers to inspect the model’s step-by-step logic and identify points of failure. Another is to strengthen attribution analysis—particularly of the vision module—using techniques such as image masking or attention map inspection to reveal which visual elements the model relies on during reasoning.

These efforts can enrich the understanding of model behavior and better inform the design of benchmarks. As such, benchmark evolution should be carefully paced. Rather than frequent, disjointed updates, a phased or tiered update cycle could synchronize better with the natural R\&D rhythm, encouraging more strategic improvements in model design and evaluation.

\appendix

\section{Details of Reasoning Tasks}
\label{Appendix:tasks}

There are nine tasks in total in our benchmark and each complexity class have three unique problem categories.

\subsection{P (Polynomial time) Tasks}

This class consists of tasks that can be solved by a deterministic Turing machine in polynomial time. Essentially, it represents tasks that are efficiently solvable. We include three P problems in the benchmark, namely Binary Search Problem (BSP), Edit Distance Problem (EDP), and Shortest Path Problem (SPP).

\subsubsection{The Binary Search Problem (BSP)}

BSP is defined by the task of locating the index of a target value within a given array as if it were sorted. The problem is presented with an unsorted array $A$ of $n$ elements and a target value $T$. The objective is to determine the index $i$ that $T$ would occupy if $A$ were sorted in ascending order. Formally, after sorting $A$, the goal is to find an index $i$ such that $A[i]=T$, or to conclude that the target is not present. This process inherently involves two stages: sorting the array and then searching for the target. The search phase is efficiently handled by algorithms like binary search, which iteratively reduces the search space and operates in logarithmic time. BSP is structured into 10 complexity levels. These levels scale by increasing the array length by one (from 3 to 12) and expanding the upper limit of the number range by 5 (from (1, 15) to (1, 60)) at each level.

\subsubsection{Edit Distance Problem (EDP)}
EDP is about finding the minimum number of operations required to transform one string into another. Given two strings, $A$ and $B$, of lengths $m$ and $n$ respectively, the aim is to determine the minimum number of operations needed to convert $A$ into $B$. The allowable operations are insertion, deletion, and substitution of a single character. Formally, the problem can be defined as finding a minimum number $d$ such that string $A$ can be transformed into string $B$ using $d$ operations. This algorithm has a time complexity of $\mathcal{O}(ab)$ where $a$ and $b$ are the lengths of the strings. When the full dynamic programming table is constructed, its space complexity is also $\mathcal{O}(ab)$. EDP has widespread applications, especially in fields like computational biology for sequence alignment, natural language processing for spell checking and correction, and in data analysis for measuring similarity between data strings. EDP is structured into 10 complexity levels, tailored to measure the minimal number of edits required to transform one string into another. Each level is characterized by two strings whose lengths are equal and progressively increase from 3 to 12 characters from Level 1 to Level 10. Concurrently, the character range for constructing these strings is expanded, starting with the first 6 letters and extending by 2 additional letters at each subsequent level.

\subsubsection{Shortest Path Problem (SPP)}
SPP is about finding the shortest path between two nodes in a non-negative weighted graph. In our experiments, we ask for the shortest path between the first and last nodes. Given a graph $G = (V, E)$ with a weight function $w: E \rightarrow \mathbb{R}$ assigning weights to edges, and two vertices $u$ and $v$ in $V$, the task is to find the path from $u$ to $v$ that minimizes the total weight. This is often solved using Dijkstra's algorithm which systematically expands the shortest path from the starting node until it reaches the target node. Formally, the problem is to find a path $P = (v_1, v_2, ..., v_k)$, where $v_1 = u$ and $v_k = v$, such that the sum of weights of consecutive edges in $P$, $\sum_{i=1}^{k-1} w(v_i, v_{i+1})$, is minimized. This problem can be used in network routing, GPS navigation systems, and logistics to find the shortest or most efficient path between two points. It helps in reducing travel time and costs in transportation and communication networks. SPP involves determining the shortest path across different complexity levels, ranging from Level 1 to Level 10. Each level is characterized by an increasing number of nodes (starting from 4 to 13), edges (5 to 14), and a maximum weight (6 to 15) that escalates linearly with each level. The complexity of the problem scales by adding one node, one edge, and increasing the maximum weight by one for each subsequent level.

\subsection{NP-complete problems}

This is a subset of NP. A problem is NP-complete if it is in NP and as hard as any problem in NP. If any NP-complete problem can be solved in polynomial time, then every problem in NP can also be solved in polynomial time. We include three NP-complete problems that are not in P in the benchmark, namely Traveling Salesman Problem Decision Version (TSP-D), Graph Coloring Problem Decision Version (GCP-D), and Knapsack Problem (KSP).

\subsubsection{Traveling Salesman Problem (Decision Version, TSP-D)}
TSP-D is concerned with determining if a salesman can complete a route, visiting each city at least once, with the total travel distance being less than a specified value. Given a complete graph $G = (V, E)$ with vertices $V$ representing cities and edges $E$ representing paths between cities, each edge $(i, j)$ is assigned a distance $d(i, j)$. The decision version of this problem asks whether there exists a tour (a sequence of cities) such that the total distance of the tour is less than or equal to a given value $D$. Formally, the problem can be stated as finding a permutation $P$ of the set of cities ${1, 2, ..., n}$ that satisfies the condition $\sum_{i=1}^{n-1} d(P(i), P(i+1)) + d(P(n), P(1)) \leq D$. This problem is useful in logistics and supply chain management in planning efficient delivery routes and schedules \cite{roberti2021exact}. TSP-D configuration spans 10 complexity levels with node counts from 4 to 13, similar to the TSP. It also introduces a threshold of 0.75, setting factor of the allowed travel distance to the total possible distance.

\subsubsection{Graph Coloring Problem (Decision Version, GCP-D)}
GCP-D involves determining if it is possible to color the vertices of a graph using a given number of colors so that no two adjacent vertices share the same color. Given an undirected graph $G = (V, E)$, with $V$ representing vertices and $E$ representing edges, the goal is to find out if there is a way to assign one of $k$ colors to each vertex such that for any edge $(u, v) \in E$, the vertices $u$ and $v$ have different colors. The formal statement is to determine if there exists a coloring function $c: V \to {1, 2, ..., k}$ such that for every edge $(u, v) \in E$, $c(u) \neq c(v)$. It has wide applications in Round-Robin Sports Scheduling, Aircraft scheduling, and Biprocessor tasks \cite{ahmed2012applications}. GCP-D has difficulty levels 1 to 10 with questions of 6, 8, 10, 12, 14, 16, 18, 20, 22, and 24 average edges and 6, 7, 8, 9, 10, 11, 12, 13, 14, and 15 nodes. Beginning with graphs of 6 nodes and 6 edges, each subsequent level incorporates an additional 2 edges and 1 node, culminating in graphs of 24 edges and 15 nodes.

\subsubsection{Knapsack Problem (KSP)}
KSP asks whether a subset of items can be chosen to fit into a knapsack of fixed capacity without exceeding it, while also maximizing the total value of the selected items. Consider a set of items, each with a weight $w_i$ and a value $v_i$, and a knapsack with a weight capacity $W$. The problem is to select a subset of these items such that the total weight does not exceed $W$ and the total value is maximized. Formally, let $x_i$ be a binary variable indicating whether item $i$ is included in the knapsack ($x_i = 1$) or not ($x_i = 0$). The problem can be stated as maximizing $\sum_{i=1}^{n} v_i x_i$ subject to the constraint $\sum_{i=1}^{n} w_i x_i \leq W$, where $n$ is the number of items. It is used in resource allocation and budgeting where the goal is to maximize the total value of a selection under a weight or cost constraint. Applications include cargo loading, and electric vehicle charging \cite{sun2020competitive, cho2019knapsack}. KSP is organized into 10 levels, each marked by an ascending number of items (from 4 to 13), a weight range (1 to the level number), a value range (identical to the weight range), and a knapsack capacity that starts at 20 and increases by 5 with each level. The complexity elevates by broadening the weight and value ranges, adding more items, and enlarging the knapsack's capacity.

\subsection{NP-hard problems}

These problems are at least as hard as the hardest problems in NP. They may not necessarily be in NP (i.e., they may not have solutions verifiable in polynomial time) but solving an NP-hard problem in polynomial time would imply that P = NP. We include three NP-hard problems that are not reducible to NP-complete problems in the benchmark, namely Traveling Salesman Problem Optimization Version (TSP), Graph Coloring Problem Optimization Version (GCP), and Meeting Scheduling Problem (MSP).

\subsubsection{Traveling Salesman Problem (Optimization Version, TSP)}
TSP-O involves finding the shortest route for a salesman to visit each city exactly once and return to the starting city. Given a complete graph $K_n$ with $n$ vertices, where each vertex represents a city and each edge $(i, j)$ is assigned a non-negative cost or distance $d(i, j)$, the problem is to find the shortest possible route that visits each city exactly once and returns to the origin city. Formally, let $P$ be a permutation of the set of cities ${1, 2, ..., n}$ representing the order in which the cities are visited. The traveling salesman problem can be formulated as finding the permutation $P$ that minimizes the total travel cost, given by the function $f(P) = d(P(n), P(1)) + \sum_{i=1}^{n-1} d(P(i), P(i+1))$. This problem is important in operational research and logistics to find the most efficient route to visit multiple locations and return to the origin, particularly route planning for delivery services, maintenance operations, and sales. TSP is structured across 10 complexity levels, each defined by a set number of nodes ranging from 4 to 13. The complexity of the problem increases with the addition of more nodes, enhancing the challenge of finding the shortest possible route that visits each node exactly once and returns to the starting point.

\subsubsection{Graph Coloring Problem (Optimization Version, GCP)}
GCP-O refers to the problem of coloring vertices of a graph in such a way that no two adjacent vertices have the same color. Given an undirected graph $G = (V, E)$, where $V$ is the set of vertices and $E$ is the set of edges, assign a color to each vertex such that no two adjacent vertices have the same color. Formally, let $c: V \to C$ be a function that assigns a color from a set of colors $C$ to each vertex in $V$. The graph coloring problem can be formulated as finding a proper coloring, i.e., a function $c$ such that for every edge $(u, v) \in E$, $c(u) \neq c(v)$. This problem is used in constraint satisfaction problems and applied in exam timetabling and register allocation in compilers \cite{lintzmayer2011register}. GCP is devised with 10 levels of complexity, featuring questions with average edges ranging from 6 to 24 and nodes from 6 to 15. Beginning with graphs of 6 nodes and 6 edges, each subsequent level incorporates an additional 2 edges and 1 node, culminating in graphs of 15 nodes and 24 edges, progressively elevating the difficulty of assigning colors to each node without any two adjacent nodes sharing the same color.

\subsubsection{Meeting Scheduling Problem (MSP)} 
MSP deals with allocating time slots for meetings such that all constraints, including participant availability and room capacity, are satisfied without overlaps. Given a set of $n$ participants and their availability for $m$ time slots, find a schedule that maximizes the number of participants who can attend the meeting. Formally, let $A = {a_1, a_2, ..., a_n}$ be the set of participants and $T = {t_1, t_2, ..., t_m}$ be the set of time slots. For each participant $a_i$, let $S_i$ be a subset of $T$ representing the times when $a_i$ is available and $m_i$ be a subset of meetings that are required to attend. The meeting scheduling problem can be formulated as finding a subset $S \subseteq T$ such that $|{a_i \in A | S_i \cap S \neq \emptyset}|$ is maximized. In other words, the aim is to find a scheduling subset $S_i$ where the collective availability of participants intersects with $S_i$, ensuring maximum participation. This problem is crucial in organizational management for scheduling meetings involving multiple participants with varying availability. It ensures optimal utilization of time and resources and is used in corporate scheduling systems and collaborative software \cite{bofill2022constraint}. MSP outlines complexity across 10 levels, determined by an increasing number of meetings (2 to 11), participants (one more than the number of meetings, i.e., 3 to 12), and time slots (two more than the number of meetings, i.e., 4 to 13). Each level advances the problem's complexity by adding one meeting, consequently increasing the number of participants and time slots required for scheduling.

\section{NPHardEval Benchmark Consistency}
\label{Appendix:consistency}

The NPHardEval Benchmark is designed for high consistency, a feature verified through preliminary assessments and supported by ongoing developments for open-source checks. For each reasoning task, such as the Binary Search Problem (BSP), the benchmark comprises three distinct versions (V0, V1, and V2) to assess the stability of a model's problem-solving capabilities. We tested this on twelve open-source models — DeepSeekVL2, Intern, Qwen2, Ovis, Llava\_Qwen, NVLM, PointsTest, Ola\_7B, Llama90BVision, GLM, Qwen25VL72B, and Gemma — as well as proprietary models including GPT-4o-2024-11-20, Gemini 2.0 Flash Exp, and Claude 3.7 Sonnet-20250219. By calculating the performance variance on each problem type across the benchmark versions, we generated 135 data points for model-specific consistency analysis. The summary statistics are as follows:
\begin{itemize}
    \item Minimum variance: 0.0 (approximately half of the variances are near zero)
    \item Mean variance: 0.01167
    \item Maximum variance: 0.11029
\end{itemize}
These statistics indicate minimal variance across the benchmark versions, suggesting a high level of consistency with each update.

\section{Recognition Accuracy}
\label{Appendix:recog}

\begin{figure*}[ht] 
    \centering
    \includegraphics[width = 1\textwidth]{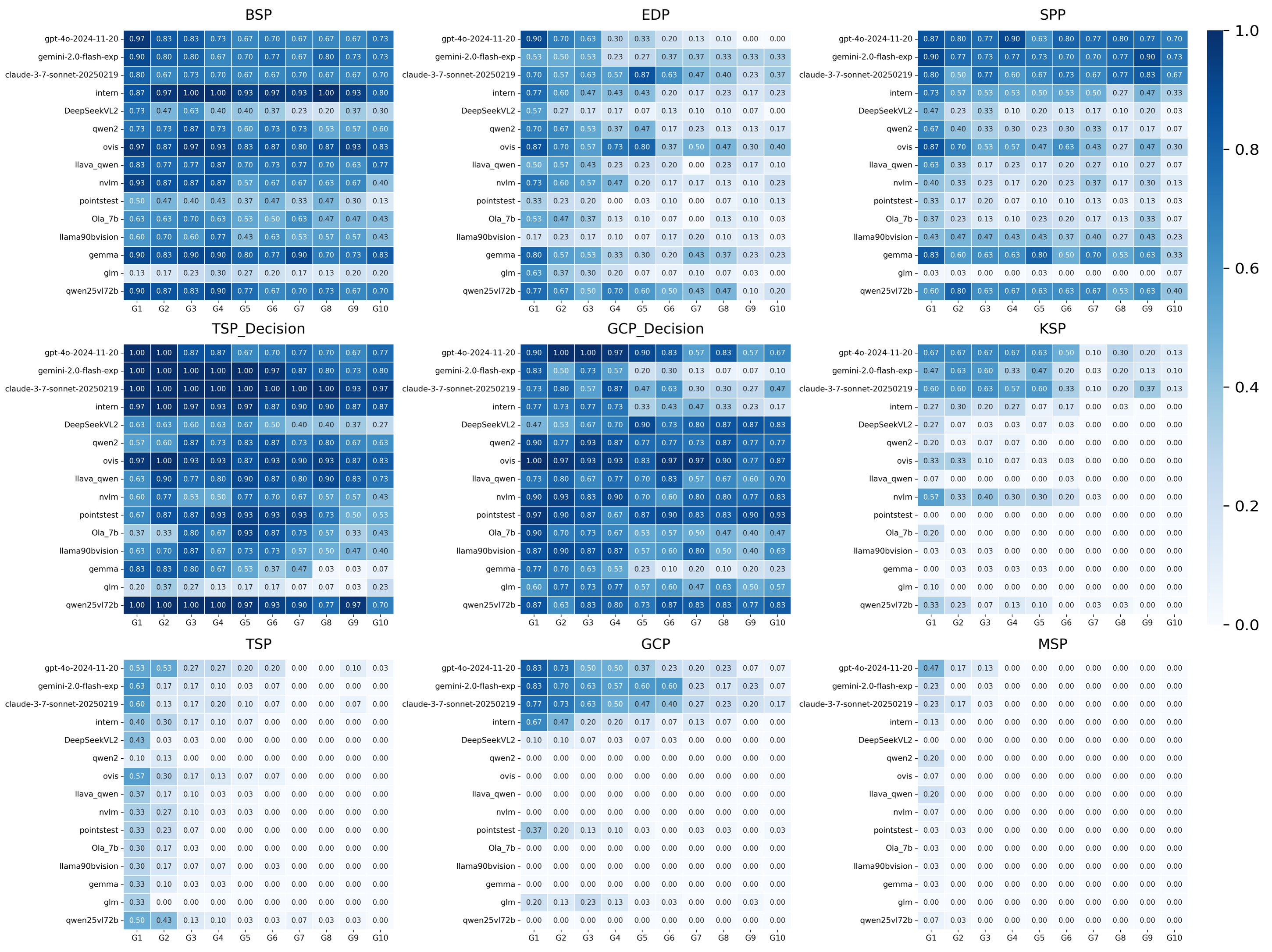}
    \vspace{-10pt}
    \caption{Prompt recognition across models, complexity levels, and difficulty levels}
    \vspace{-10pt}
    \label{Fig:RA_heatmap}
\end{figure*}

Recognition is a critical initial step, as a model's ability to understand the prompt directly impacts its performance on downstream tasks. As shown in Figure 9, closed-source models like gpt-4o-2024-11-20, gemini-2.0-flash-exp, and claude-3-7-sonnet-20250219 generally exhibit superior and more consistent recognition accuracy across the majority of tasks. It is also notable that performance varies dramatically between different task types; for example, most models excel at TSP\_Decision while struggling significantly with TSP, indicating that task formulation is a key factor in recognition success.

Despite this trend, open-source models demonstrate impressive recognition capabilities, matching or exceeding their closed-source counterparts on specific tasks. For instance, Intern achieves perfect recognition (1.0) on several difficulty levels of the BSP task. Similarly, models like ovis, nvlm, and qwen25vl72b show outstanding performance, particularly on the TSP\_Decision and GCP\_Decision tasks, often maintaining near-perfect accuracy. This not only highlights the significant variation in prompt recognition ability across different tasks but also underscores the unique specializations of individual LVLMs, likely stemming from differences in their training datasets and architectural designs.

\bibliographystyle{ACM-Reference-Format}
\bibliography{references}



\end{document}